# A MEASURE-FREE APPROACH TO CONDITIONING

I.R. Goodman

Command & Control Department
Code 421
NAVAL OCEAN SYSTEMS CENTER
San Diego, California 92152

## ABSTRACT

In an earlier paper, a new theory of measure-free "conditional" objects was presented. In this paper, emphasis is placed upon the motivation of the theory. The central part of this motivation is established through an example involving a knowledge-based system. In order to evaluate combination of evidence for this system, using observed data, auxiliary attribute and diagnosis variables, and inference rules connecting them, one must first choose an appropriate algebraic logic description pair (ALDP): a formal language or syntax followed by a compatible logic or semantic evaluation (or model). Three common choices- for this highly non-unique choice - are briefly discussed, the logics being Classical Logic, Fuzzy Logic, and Probability Logic. In all three, the key operator representing implication for the inference rules is interpreted as the often-used disjunction of a negation $(b \ni a) = (b' \vee a)$, for any events $a,b$.

However, another reasonable interpretation of the implication operator is through the familiar form of probabilistic conditioning. But, it can be shown - quite surprisingly - that the ALDP corresponding to Probability Logic cannot be used as a rigorous basis for this interpretation! To fill this gap, a new ALDP is constructed consisting of "conditional objects", extending ordinary Probability Logic, and compatible with the desired conditional probability interpretation of inference rules. It is shown also that this choice of ALDP leads to feasible computations for the combination of evidence evaluation in the example. In addition, a number of basic properties of conditional objects and the resulting Conditional Probability Logic are given, including a characterization property and a developed calculus of relations.

## 1. INTRODUCTION

This paper is complementary to a previous one [1] in which measure-free conditional objects are first introduced. In that paper, emphasis was placed upon a summary of the various mathematical properties that are derivable. In this paper, motivation for the use of conditional objects is underscored, followed by a brief overview of results. A more thorough presentation, together with all relevant proofs, can be found in [2].

The basic questions that are demanded of a new theory include:

> What use is it ? Is it necessary ?
> Does it solve an existing problem ?
> Is it truly novel ?
> Does it tie-in with past literature in the field ?
> Is it mathematically sound and sufficiently rich to lead to further deeper results and applications?

It is the hope of this paper and accompanying work to provide positive answers to the above questions through the development of conditional object theory.

In a typical knowledge-based system, a collection of inference rules is present, each rule connecting potential observed data through auxiliary attributes to potential parameter estimates or diagnoses. Each rule also has, as a main connector, some form of implication. Thus, in evaluating such systems, it is critical that consistent and feasible interpretations and computations be made for these operators.

At present, there is no sound logic of conditional events, analogous to ordinary Probability Logic, in use. Thus no systematic approach exists for combination of evidence problems, when individual inference rules are interpreted through conditional probabilities. Indeed, D. Lewis [9] pointed out in 1976 that one could not identify implication with conditioning in the probability sense. That is, if

$$(b \ni a) = (a|b) \in \Omega ,  \qquad (1.1)$$

where $\Omega$ is some fixed boolean algebra of events or propositions $a,b,...$, then formally applying a given probability measure $p:\Omega \to [0,1]$ to both sides yields

$$p(b \ni a) = p((a|b)) = p(a|b) \stackrel{d}{=} p(a \cdot b)/p(b) , \qquad (1.2)$$

provided that $p(b) > 0$. But, if one makes the common identification (but by no means, the only possible)

$$(b \ni a) = (b \to a) \stackrel{d}{=} (b' \vee a) , \qquad (1.3)$$

then one can show, by use of elementary properties of conditional and unconditional probabilities, that

$$p(b \to a) = p(a|b) + p(b') \cdot p(a'|b)$$
$$\geq p(a|b)$$
$$\geq p(a \cdot b) , \qquad (1.4)$$

with strict inequality holding in general. (In fact, it is rather easy to construct examples where $p(b \to a)$ is close to unity while $p(a|b)$ is close to zero.) Furthermore, Calabrese [10],[11] has shown that not only will $\to$ not work in (1.2), but no boolean function of two arguments possibly representing implication, will satisfy (1.2).

Yet, often individuals assume the identification in (1.1) - at least tacitly - and manipulate and interchange conditional probabilities and implications, noting an easily-derivable calculus of relations for such implications. (See, e.g., Table 1.) In fact, Stalnaker's Thesis [8] carries out this identification; but see Lewis' criticism [9].

Thus, we must pose the basic question: Can we make sense of "conditional object" (a|b) compatible with conditional probability p(a|b) ? Also, how do we compute (a|b) v (c|d) and in turn evaluate the expression p((a|b)v(c|d)) ? Lastly, can we use such entities in combination of evidence problems in conjunction with knowledge-based systems?

Another approach to avoiding the establishment of conditional objects, in effect, is to equate a given collection of conditional probabilities with corresponding common antecedent conditional probabilities, through formation of appropriate joint events. But this approach can also be shown to lead to certain difficulties conceptually as well as computationally [2].

Additional discussions concerning Lewis' "triviality result" concerning Stalnaker's Thesis can be found in [12],[13],[14].

The proposed remedy to the above problem involves an extension of coset theory as applied to boolean algebras, where the original boolean algebra of events $\Omega$ is replaced by the union of all principal ideal quotient rings of $\Omega$. The fundamental justification for this will be given below, followed by an example illustrating how conditional objects and Conditional Probability Logic can be directly utilized in a knowledge-based system. (See Section 2.)

However, let us first back up and consider how a typical combination of evidence problem can be perceived. Figure 1 illustrates the basic information processing flow from the inception of the problem to the decision process. This processing consists of five subdivisions in sequence:

1. Cognition: Initial processing of information.

2. Natural Language Formulation: Relevant to all narratives produced by human observers.

3. Primitive Symbolic Formulation of Information: Formation of well-defined formulas or strings of information, without refined constraints. Use of basic formal connects for:&, · ; or, v ; not, ( )'; implication, $\Rightarrow$ .

4. Full Formal Language: Use of all relevant syntax, constraints on connectors, such as associativity, commutativity, absorbing, distributivity, demorgan,etc.

5. Full Semantic Evaluations or Logic : Must be in model form,i.e., some type of homomorphism preserving basic structure of full formal language.

It is the choice of the last two subprocesses with which we are concerned here. We will call such a pair of subprocesses 4 and 5 in Figure 1 an <u>algebraic logic description pair</u> (ALDP). As given in Figure 1,let:

ALDP 1 = (boolean algebra $\Omega$, Classical Logic (CL))

ALDP 2 = (modified boolean algebra $\Omega_0$,Fuzzy Logic(FL))

ALDP 3 = (boolean algebra $\Omega$, Probability Logic(PL)).

In all three ALDP's above, implication, from now on, is to be interpreted only via (1.3).

See [3], Chapter 2 for general background concerning formal language and semantic evaluation in modeling knowledge-based systems. See also [4] for an excellent survey of multivalued logics, including PL. See [5] for FL and [6] for boolean algebras and rings. Future efforts will deal with extensions of these ideas to nonmonotonic logics as presented,e.g., in [7].

For purposes of completeness, let us next briefly review each ALDP, presenting an abridged calculus of operations involving implication and semantic evaluation for use in the ensuing example in Section 2.

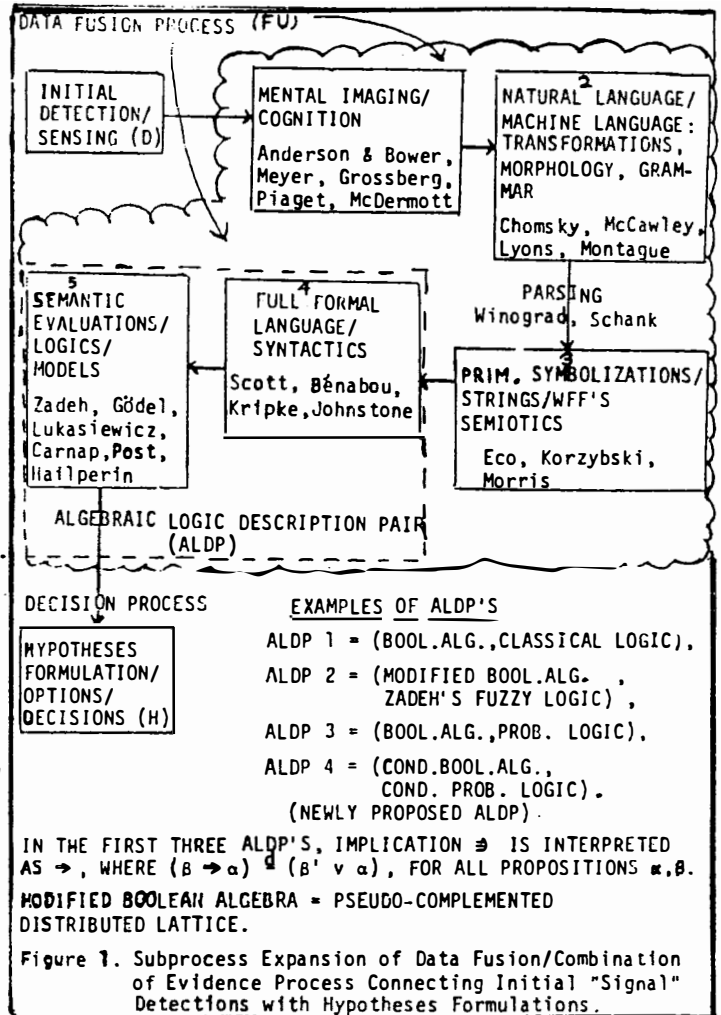

Figure 1. Subprocess Expansion of Data Fusion/Combination of Evidence Process Connecting Initial "Signal" Detections with Hypotheses Formulations.

EXAMPLES OF ALDP'S
ALDP 1 = (BOOL.ALG.,CLASSICAL LOGIC),
ALDP 2 = (MODIFIED BOOL.ALG. , ZADEH'S FUZZY LOGIC) ,
ALDP 3 = (BOOL.ALG.,PROB. LOGIC),
ALDP 4 = (COND.BOOL.ALG., COND. PROB. LOGIC).
(NEWLY PROPOSED ALDP).
IN THE FIRST THREE ALDP'S, IMPLICATION $\Rightarrow$ IS INTERPRETED AS $\rightarrow$, WHERE $(\beta \rightarrow \alpha) \overset{d}{=} (\beta' v \alpha)$, FOR ALL PROPOSITIONS $\alpha, \beta$.
MODIFIED BOOLEAN ALGEBRA = PSEUDO-COMPLEMENTED DISTRIBUTED LATTICE.

Given boolean algebra $\Omega$, for all $a,b,c,a_i,b_i \in \Omega$, $i=1,..,m$ :

$$(b \Rightarrow a) = (b \Rightarrow a \cdot b) \in \Omega , \quad (1.5)$$

$$(1 \Rightarrow a) = a , \quad (1.6)$$

(chaining) $\quad (c \Rightarrow a \cdot b) = (c \Rightarrow b) \cdot (b \cdot c \Rightarrow a), \quad (1.7)$

$$(b \Rightarrow a)' = a' \cdot b \neq (b \Rightarrow a') , \text{ in general}(1.8)$$

$$\overset{m}{\underset{i=1}{v}}(b_i \Rightarrow a_i) = (\overset{m}{\underset{i=1}{\cdot}} b_i \Rightarrow \overset{m}{\underset{i=1}{v}} a_i), \quad (1.9)$$

$$\overset{m}{\underset{i=1}{\cdot}}(b_i \Rightarrow a_i) = ((\overset{m}{\underset{i=1}{v}} a_i' \cdot b_i \text{ v } \overset{m}{\underset{i=1}{\cdot}} b_i) \Rightarrow \overset{m}{\underset{i=1}{\cdot}} a_i) (1.10)$$

Homomorphisms hold for $b_1 = \cdots = b_m = b$ :

$$\overset{m}{\underset{i=1}{v}} (b \Rightarrow a_i) = (b \Rightarrow \overset{m}{\underset{i=1}{v}} a_i), \quad (1.11)$$

$$\overset{m}{\underset{i=1}{\cdot}} (b \Rightarrow a_i) = (b \Rightarrow \overset{m}{\underset{i=1}{\cdot}} a_i), \quad (1.12)$$

but $\overset{m}{\underset{i=1}{+}} (b \Rightarrow a_i) = \begin{cases} (b \Rightarrow \overset{m}{\underset{i=1}{+}} a_i) , \text{ if m is odd} \\ \overset{m}{\underset{i=1}{+}} a_i , \text{ if m is even} \end{cases} \quad (1.13)$

Table 1. ALDP 1: Calculus of Operations for Implication (1.3).

Given boolean algebra $\Omega$ and semantic evaluation

$$\|\ \|:\Omega \to \{0,1\}\ ,\ \|0\|=0\ ,\ \|1\|=1, \quad (1.14)$$

for all $a,b \in \Omega$:

$$\|a \lor b\| = MAX(\|a\|,\|b\|), \quad (1.15)$$
$$\|a \cdot b\| = MIN(\|a\|,\|b\|), \quad (1.16)$$
$$\|a'\| = 1 - \|a\|, \quad (1.17)$$
$$\|a+b\| = \|a' \cdot b \lor a \cdot b'\|, \quad (1.18)$$
$$\|b \to a\| = MAX(1 - \|b\|, \|a\|). \quad (1.19)$$

Table 2. ALDP 1 : CL Evaluation for $\Omega$.

---

Given modified boolean algebra $\Omega_0$, for all $a,b,c,a_i,b_i$, $i=1,\ldots,m$:

$$(b \to a) = (b \to a \cdot b) \in \Omega_0, \quad (1.20)$$

if $a \leq b$, i.e., $a = a \cdot b$, for middle equality to hold.

$$(1 \to a) = a, \quad (1.21)$$

(chaining) $(c \to (a \cdot b \lor b \cdot b')) = (c \to b) \cdot (b \cdot c \to a),$ (1.22)

$$(b \to a)' = a' \cdot b \neq (b \to a'), \text{ in general},(1.23)$$

$$\bigvee_{i=1}^{m}(b_i \to a_i) = (\cdot \underset{i=1}{\overset{m}{\cdot}} b_i \to \underset{i=1}{\overset{m}{\vee}} a_i), \quad (1.24)$$

$$\underset{i=1}{\overset{m}{\cdot}}(b_i \to a_i) = ((\underset{i=1}{\overset{m}{\vee}} a_i' \cdot b_i \lor \underset{i=1}{\overset{m}{\cdot}} b_i) \to \underset{i=1}{\overset{m}{\cdot}} a_i). \quad (1.25)$$

Homomorphisms hold for $b_1 = \cdots = b_m = b$:

$$\underset{i=1}{\overset{m}{\vee}}(b \to a_i) = (b \to \underset{i=1}{\overset{m}{\vee}} a_i), \quad (1.26)$$

$$\underset{i=1}{\overset{m}{\cdot}}(b \to a_i) = (b \to \underset{i=1}{\overset{m}{\cdot}} a_i), \quad (1.27)$$

but $\underset{i=1}{\overset{m}{+}}(b \to a_i) = ((b \lor b') \to \underset{i=1}{\overset{m}{+}} a_i).$ (1.28)

Table 3. ALDP 2 : Calculus of Operations for Implication (1.3).

---

For ALDP 2, for FL evaluation over $\Omega_0$, where now "poss" indicates Zadeh's possibility or membership function

$$\|\ \| = poss : \Omega_0 \to [0,1]\ ,\ \|0\|=0, \|1\|=1, \quad (1.29)$$

replacing (1.14), see Table 2 for all evaluations.

For ALDP 3, formal language is same ($\Omega$ boolean) as for ALDP 1; thus see Table 1, for calculus of operations for implication.

For ALDP 3, the semantic evaluation becomes the standard probability type as given in Table 4:

---

Given boolean algebra $\Omega$ and semantic evaluation

$$\|\ \| = p:\Omega \to [0,1]\ ,\ p(0)=0\ ,\ p(1) = 1, \quad (1.30)$$

for any finitely additive probability measure $p$, for all $a,b,a_i \in \Omega$, $i=1,\ldots,m$:

$$p(a \lor b) = p(a) + p(b) - p(a \cdot b), \quad (1.31)$$

$$p(\underset{i=1}{\overset{m}{\vee}} a_i) = \sum_{\emptyset \neq J \subseteq \{1,\ldots,m\}} (-1)^{card(J)+1} \cdot p(\underset{i \in J}{\cdot} a_i), \quad (1.32)$$

$$p(a') = 1 - p(a), \quad (1.33)$$
$$p(b \to a) = p(b') + p(a \cdot b). \quad (1.34)$$

Table 4. ALDP 3: PL Evaluation for $\Omega$.

---

## 2. AN EXAMPLE

With all the preliminaries out of the way, consider next a simple medical diagnosis system as illustrated in Figure 2 (next page). The basic implementation scheme in combining evidence is given below, all corresponding in Figure 1 to information processing, up to and including Stage 3, Primitive Symbolizations, prior to choice of ALDP, making up the last two stages (4,5):

1. Choose data event variable set G from attribute variables. Observe symptoms $y \in dom(G)$.

2. Form $I_y$, the set of all inference rules $\psi$ where in either antecedent and/or consequent some variables in G appear. $I_y$ can also be considered the potential "firing" class of inference rules, if CL and ALDP 1 were chosen.

3. Form

$$f(G[y],A_y[W],\Theta_y[W_2]) = y \cdot (\underset{\psi \in I_y}{\cdot} \psi[W]), \quad (2.1)$$

the conjunction of all data with relevant inference rules, where we denote $\Theta_y$ as the set of all diagnoses variables $\Theta_i \in I_y$; $A_y$ is the set of all attribute variables in $I_y \to \Theta_y$; and $W=(W_1,W_2)$ represents the domain variables, with $W_2$ corresponding to $\Theta_y$.

4. Compute

$$(y \gg \Theta_y[W_2]) \overset{d}{=} \underset{W_1 \in dom(A_y)}{\vee} ( f(G[y],A_y[W],\Theta_y[W_2]) ), \quad (2.2)$$

the full "integrated-out" form representing the posterior relation between symptoms and diagnoses.

Next, as a particularization, suppose now that in terms of the above scheme attribute, $b_1$ is selected and $(106°, REDDISH)$ is observed. Thus, using Figure 2:

$$\begin{cases} G=\{b_1\}\ ;\ y=b_1[(106°,REDDISH)]\ ; \\ I_y = \{(y \ni a_1),((y \lor b_2) \ni (a_2 \lor a_3)),(b_2 \ni \Theta_1),((a_1 \lor a_2) \ni \Theta_1)\} \\ A_y = \{a_1,a_2,a_3,b_2\}\ ;\ \Theta_y = \{\Theta_1\}\ ; \\ W = (W_1,W_2)\ ;\ W_1 = (x_1,x_2,x_3,z_2)\ ;\ W_2 = t_1\ ; \\ f(G[y],A_y[W_1],\Theta_y[W_2]) = y \cdot (y \ni a_1[x_1]) \cdot \\ \cdot ((y \lor b_2[x_2]) \ni (a_2[x_2] \lor a_3[x_3])) \cdot (b_2[z_2] \ni \Theta_1[t_1]) \cdot \\ \cdot ((a_1[x_1] \lor a_2[x_2]) \ni \Theta_1[t_1])\ ; \\ (y \gg \Theta_1[t_1]) = \underset{\begin{subarray}{c} x_i \in DOM(a_i), \\ i=1,2,3\ ; \\ z_2 \in DOM(b_2) \end{subarray}}{\vee} ( f(b_1[y],A_y[W_1],\Theta_1[t_1]) ). \end{cases} \quad (2.3)$$

The final step in the evaluation of the medical diagnosis is to choose an ALDP and apply this to (2.2) to obtain the semantic evaluation of the relation between symptoms and diagnoses. Consider, then this evaluation for the particular case given above in (2.3) for ALDP's 1,2,3.

For the formal language for ALDP 1 and 3, $\Omega$ boolean, using either Table 1 or basic properties of boolean algebras, one readily obtains

$$f(G[y],A_y[W_1],\Theta_y[W_2]) = ((\eta \cdot \Theta_1 \cdot y)' \ni (\eta \cdot \Theta_1 \cdot y)) = \eta \cdot \Theta_1 \cdot y, \quad (2.4)$$

where

$$\eta \overset{d}{=} a_1 \cdot (a_2 \lor a_3) = \eta[W_1]. \quad (2.5)$$

Thus,

$$(y \gg \Theta_1[t_1]) = \eta_0 \cdot \Theta_1[t_1] \cdot y, \quad (2.6)$$

272

| PRIMITIVE SYMBOLIC FORMULATION OF INFORMATION: EVENT VARIABLES $a_1,a_2,\ldots,b_1,b_2,\ldots,\theta_1,\theta_2$ WITH REALIZATIONS IN $\Omega$; CONNECTORS: $\cdot,v,()',\emptyset,\ldots$ | | | TYPICAL EVENTS IN $n_i$ THROUGH SUBSTITU. VAR. | DOMAIN OF VALUES |
|---|---|---|---|---|
| EVENT VARIABLES | AUXILIARY ATTRIBUTE | $a_1$=STATE OF RENAL SYSTEM | $x_1$  $a_1[x_1]$ | DOM($a_1$)={1,...,7} |
| | | $a_2$=STATE OF PULMINARY SYSTEM | $x_2$  $a_2[x_2]$ | DOM($a_2$)={1,...,18} |
| | | $a_3$=STATE OF VASCULAR SYSTEM | $x_3$  $a_3[x_3]$ | DOM($a_3$)={1,...,39} |
| | | $a_4$=STATE OF CIRCULATORY SYSTEM | $x_4$  $a_4[x_4]$ | DOM($a_4$)={1,...,9} |
| | DATA ATTRIBUTE | $b_1$=(BODY TEMP., SKIN-TONE) | $z_1$  $b_1[z_1]$ | DOM($b_1$)={(90°,91°,...,110°);(DRY,DULL,...)} |
| | | $b_2$=DEGREE OF LUNG CONGESTION | $z_2$  $b_2[z_2]$ | DOM($b_2$)={1.3,2,3.5,...,70} |
| | | $b_3$=BLOOD PRESSURE | $z_3$  $b_3[z_3]$ | DOM($b_3$)={NONE,MODER..HIGH} |
| | DIAGNOSES | $\theta_1$=DEGREE OF DISEASE STATE 1 | $t_1$  $\theta_1[t_1]$ | DOM($\theta_1$)={NONE,SOME,PROG.} |
| | | $\theta_2$=DEGREE OF DISEASE STATE 2 | $t_2$  $\theta_2[t_2]$ | DOM($\theta_2$)={NONE,MEDIUM,HIGH} |
| | | $\theta_3$=GENERAL HEALTH LEVEL | $t_3$  $\theta_3[t_3]$ | DOM($\theta_3$)={POOR,FAIR,GOOD} |

| # = CORE OF EXPERT-ESTABLISHED INFERENCE RULE FUNCTIONALS $\ast=(a\ni a)$ | ACTUAL INFERENCE RULES OBTAINED BY SUBSTITUTION OF VARIABLES |
|---|---|
| $(b_1\ni a_1) \cdot ((b_1 v b_2)\ni(a_2 v a_3))$ | $(b_1[z_1]\ni a_1[x_1])\cdot((b_1[z_1] v b_2[z_2])\ni(a_2[x_2] v a_3[x_3]))$ |
| $(b_2\ni \theta_1) \cdot ((a_1 v a_2)\ni \theta_1)$ | $(b_2[z_2]\ni \theta_1[t_1])\cdot((a_1[x_1] v a_2[x_2])\ni \theta_1)$ |
| $(a_4\ni b_3) \cdot ((b_5\cdot b_3)\ni(a_4 v a_5))$ | $(a_4[x_4]\ni b_3[z_3])\cdot((b_5[z_5]\cdot b_3[z_3])\ni(a_4[x_4] v a_5[x_5]))$ |

Figure 2. A Simple Medical Diagnosis System : Basic Structure Prior to Choice of ALDP

where (2.7)
$$n_0 \stackrel{d}{=} \underset{x_1 \in DOM(a_1)}{v} a_1[x_1] \cdot (\underset{x_2 \in DOM(a_2)}{v} a_2[x_2] v \underset{x_3 \in DOM(a_3)}{v} a_3[x_3])$$

Semantic evaluation (CL) for ALDP 1 is, using Table 2,

$$\|y \twoheadrightarrow \theta_1[t_1]\| = MIN(\|n_0\|, \|\theta_1[t_1]\|, \|y\|), \quad (2.8)$$

where
$$\|n_0\| = MIN(\underset{x_1 \in DOM(a_1)}{MAX} \|a_1[x_1]\|, \underset{x_i \in DOM(a_i), i=2,3}{MAX} \|a_i[x_i]\|), \quad (2.9)$$

which has obvious interpretations in CL.

On the other hand, semantic evaluation (PL) for ALDP 3, for some appropriate probability measure p is

$$p(y \twoheadrightarrow \theta_1[t_1]) = p(n_0 \cdot \theta_1[t_1] \cdot y), \quad (2.10)$$

which can be further evaluated using the expansion (1.32).

For the formal language for ALDP 2, $\Omega_0$ modified boolean, using either Table 3 or basic properties ([5], pp. 14-16), it follows that

$$f(G[y], A_y[W_1], \theta_y[W_2]) = y \cdot ((b_2' v a_2 v a_3) \cdot y' v \eta) \cdot (\theta_1 v a_1' \cdot a_2' \cdot b_2'). \quad (2.11)$$

Then, using Table 2 (see (1.29)), the semantic evaluation becomes, assuming for simplicity $\|a_1[x_1]\|$ monotone in $x_1 \in DOM(a_1)$ and assuming (2.12)

$$\underset{z_2 \in DOM(b_2)}{MIN} poss(b_2[z_2])=0= \underset{x_2 \in DOM(a_2)}{MIN} poss(a_2[x_2]); \underset{x_3 \in DOM(a_3)}{MAX} poss(a_3[x_3])=1,$$

then
$$\|y \twoheadrightarrow \theta_1[t_1]\|= \underset{x_1 \in DOM(a_1)}{MAX}(MIN(y, MAX(\|y'\|, \|a_1[x_1]\|), MAX(\|\theta_1\|, \|a_1[x_1]\|)))$$

$$= MIN(\|y\|, MAX(1-\|y\|, \|\theta_1[t_1]\|, \tfrac{1}{2})). \quad (2.13)$$

Thus if we interpret implication as in (1.3), the above all show that feasible computations can be obtained for the evaluation of the posterior relation between symptoms and diagnoses for ALDP 1,2,3.

On the other hand, a basic interpretation of implication is through probabilistic conditioning. But in light of the remarks in Section 1, if we are to have (1.2) hold, we cannot have $(b \ni a) \in \Omega$, the given boolean algebra of events for the probability space. But also following the guidelines given in Section 1 (Figure 1), we seek an ALDP, say ALDP 4, which is compatible with (1.2) and yields, hopefully, computations, no more complex than the three standard ALDP's considered for this example as a case in point. For the time being, assume ALDP 4 exists, where the calculus of operations involving conditioning interpreted as implication is given in Section 3. Then applying these results directly to (2.3) yields first (using (3.7)), after simplifying:

$$f(G[y], A_y[W_1], \theta_y[W_2]) = (n \cdot \theta_1[t_1] \cdot y | (n \cdot \theta_1[t_1] \cdot y)' v b_2[z_2]), (2.14)$$

where n is given in (2.5). In turn, using (3.6) in (2.3)

$$(y \twoheadrightarrow \theta_1[t_1]) = (n_0 \cdot \theta_1[t_1] \cdot y | (n_0 \cdot \theta_1[t_1] \cdot y)' v \tau_0), \quad (2.15)$$

where $n_0$ is given in (2.7) and

$$\tau_0 \stackrel{d}{=} \underset{z_2 \in DOM(b_2)}{v} b_2[z_2] . \quad (2.16)$$

Then, applying selected probability measure p (see (4.23))

$$\|y \twoheadrightarrow \theta_1[t_1]\| = p(y \twoheadrightarrow \theta_1[t_1]) = p(n_0 \cdot \theta_1[t_1] \cdot y | (n_0 \cdot \theta_1[t_1] \cdot y)' v \tau_0)$$

$$= p(n_0 \cdot \tau_0 \cdot \theta_1[t_1] \cdot y)/p(n_0 \cdot \theta_1[t_1] \cdot y)' v \tau_0), \quad (2.17)$$

which can be further evaluated through use of (1.32).

Note that the computations for ALDP 4 parallel closely those of ALDP 3, except for denominator term. It follows also, by direct comparisons with the above results, that indeed use of ALDP 4 results in calculations no more complicated than those for ALDP's 1,2,3.

Note also that the calculus of operations for conditional objects (again, see Section 3) is analogous to those for ALDP's 1,2,3, using (1.3) for implication, by inspection of Tables 1 and 3. Note further, that these ALDP's are not quite homomorphisms for fixed common antecedents, due to problems with negation and/or disjoint sum (+) (see (1.11)-(1.13), (1.26)-(1.28)). But, in effect, conditional probability forms hold which are natural counterparts of homomorphic relations. $\ast = v, \ast, +$:

$$p(a \ast c|b)=p((a|b) \ast (c|b)), \quad p((a|b)')=1-p(a|b)=p(a'|b), \quad (2.18)$$
Note also, $p(a|b) = p(a \cdot b|b) \quad (2.19)$

Indeed, the following result shows that if we remove the particular probability measure p in (2.18) and (2.19), prior to any evaluation, the resulting entities - conditional objects - are uniquely determined:

Theorem 2.1 Characterization of Conditional Objects.

Given a boolean algebra $\Omega$ (or equivalently, ring when considering + and $\cdot$), there is a unique space $\tilde{\Omega}$ of smallest possible classes of elements - according to subset partial ordering - denoted as <u>conditional objects</u> $(a|b), (c|d), \ldots$, for all $a, b, c, d, \ldots \in \Omega$, such that the measure-free counterparts of (2.18) (see (3.1) and (3.25)-(3.27)) and (2.19) hold. For the latter,

$$(a|b) = (a \cdot b|b), \text{ for all } a, b \in \Omega . \quad (2.20)$$

Moreover, the conditional objects constituting $\tilde{\Omega}$ coincide with all possible principal ideal cosets of ring $\Omega$, where explicitly, for all $a, b \in \Omega$,

$$(a|b)=\Omega \cdot b' + a = \Omega \cdot b' + a \cdot b = \Omega \cdot b' v a \cdot b$$

$$= \{x \cdot b' + a \cdot b | x \in \Omega\} \subseteq \Omega , \quad (2.21)$$

the principal ideal coset generated by b' with residue a (or, equivalently, $a \cdot b$).

Proof:
Use first the basic homomorphism theorem characterizing quotient rings, for (2.18); then apply equivalence class property of cosets. See [2] for details. □

With the rigorous basis for conditional objects justified above, define ALDP 4 as the pair $(\tilde{\Omega}, p)$, where $p: \Omega \to [0,1]$ is extended to $p: \tilde{\Omega} \to [0,1]$ via (1.2) or (4.23). Note also the immediate relation from (2.21):

$$(a|b)=(c|d) \text{ iff } a \cdot b=c \cdot d \text{ and } b=d ; \text{all } a,b,c,d \in \Omega. (2.22)$$

In addition, since for all $a \in \Omega$,
$(a|1) = a ,$ (2.23)
then $\tilde{\Omega}$ extends $\Omega$: $\Omega \subseteq \tilde{\Omega}.$ (2.24)

In the approach taken here, all results involving conditional objects are derived from first principles. In this vein, define all operations among con-ditional objects as the natural class or component-wise extensions of the corresponding operations among the elements. Thus for example, for $a,b,c,d \in \Omega$,
$$(a|b)\cdot(c|d) \stackrel{d}{=} \{q\cdot r | q\in(a|b), r\in(c|d)\} = \{(x\cdot b'+a)(y\cdot d'+c)|x,y\in\Omega\}, \quad (2.25)$$
also a subset of $\Omega$ as are $(a|b)$ and $(c|d)$. From this, it follows immediately that conditioning as defined here is essentially the functional inverse of one-sided conjunction, i.e., the following hold for all $a,b \in \Omega$:
$$(a|b)\cdot b = a\cdot b \; ; \; (a|b) = \{x | x \in \Omega, x\cdot b = a\cdot b\}. \quad (2.26)$$

Hailperin [15] considered conditional objects, extending some of Boole's original ideas, but avoided combining these entities when antecedents differ-through use of universal algebras and partially-defined operators. Domotor [16], following the direction of "qualitative probability structures, as used in subjective probability theory and preference orderings, developed a rather cumbersome indirect approach, not realizing the rich structure of $\tilde{\Omega}$. (See, e.g., Theorems 3.2,3.3 in this paper.) Nute [14], among others[13], has also considered "conditional logics", which appear to be generally related to this work, but differ considerably in structure. Much work remains in tying-in these concepts with conditional objects as envisioned here. Finally, the pioneering work of Calabrese [11] must be mentioned as the direct cause of the current work. Although his definition for conditional objects can be shown to be equivalent ([1],(2.19)-(2.25)), Calabrese proposes ad hoc definitions for operators upon them, in contradistinction to the first principles approach taken here.

## 3. BASIC PROPERTIES OF CONDITIONAL OBJECTS

### Theorem 3.1

The boolean operations $+$, $\vee$, $\cdot$, $(\;)'$ are all well-defined over $\tilde{\Omega}$ as the natural class extensions of the ordinary counterparts over $\Omega$. Indeed:
$$(a|b)' = (a'|b) = (a'\cdot b|b), \quad (3.1)$$
$$(a|b) + (c|d) = (a+c|bd) = (ab+cd|bd), \quad (3.2)$$
$$(a|b) \vee (c|d) = (a \vee c | ab \vee cd \vee bd)$$
$$= (ab \vee cd | ab \vee cd \vee bd), \quad (3.3)$$
$$(a|b) \cdot (c|d) = (a\cdot c | a'b \vee c'd \vee bd)$$
$$= (abcd | a'b \vee c'd \vee bd). \quad (3.4)$$

$+$, $\vee$, $\cdot$ are all associative and commutative over $\tilde{\Omega}$, and hence extendable unambiguously to any number of arguments. Specifically, for any $n \geq 1$ and any $a_i, b_i \in \Omega$, $i=1,...,n$
$$(a_1|b_1) + \cdots + (a_m|b_m) = (a_1 + \cdots + a_m | b_1 \cdots b_m), \quad (3.5)$$
$$(a_1|b_1) \vee \cdots \vee (a_m|b_m) = (a_1 \vee \cdots \vee a_m | a_1\cdot b_1 \vee \cdots \vee a_m\cdot b_m \vee b_1\cdots b_m), \quad (3.6)$$
$$(a_1|b_1)\cdots(a_m|b_m) = (a_1\cdots a_m | a_1'\cdot b_1 \vee \cdots \vee a_m'\cdot b_m \vee b_1\cdots b_m). \quad (3.7)$$

### Proofs and remarks.

An outline of the algebraic nature of the proof is given here. First, recall a ring is boolean iff it is idempotent - $a^2=a$, for all $a$ in ring. More generally, $\Omega$ is Von Neumann regular [17] iff for all $a \in \Omega$, there exists $\lambda_a \in \Omega$ with $a = \lambda_a \cdot a^2$, assuming commutativity with unity.

Note first that for any commutative ring with unity, say $\Omega$, and ideals $I, J \subseteq \Omega$ and $a, c \in \Omega$,
$$(I+a) + (J+c) = (I+J) + (a+c), \quad (3.8)$$
where $I+J$ is also an ideal of $\Omega$. In particular, define for any $a, b \in \Omega$ (Von Neumann) regular,
$$a' \stackrel{d}{=} 1 - \lambda_a \cdot a \; , \; a \vee b \stackrel{d}{=} \lambda_a \cdot a + \lambda_b \cdot b - \lambda_a \cdot a \cdot \lambda_b \cdot b, \quad (3.9)$$
noting the reduction to the boolean case, i.e., when $\lambda_a = \lambda_b = 1$, where the relations simplify to
$$a' = 1-a = 1+a \; , \; a \vee b = a+b-a\cdot b = a+b+a\cdot b. \quad (3.10)$$
Then letting $I = \Omega\cdot b'$ and $J = \Omega\cdot d'$ for $\Omega$ regular, it follows that (3.8) becomes

$$(\Omega\cdot b' + a) + (\Omega\cdot d' + c) = \Omega\cdot(b'\vee d') + (a+c)$$
$$= \Omega\cdot(b\cdot d)' + (a+c), \quad (3.10a)$$
by DeMorgan's relation valid for regular rings and the property of regular rings whereby sums of principal ideals are principal ideals computable as above. Thus using (2.20) and (2.21), (3.2) holds.

Next, using (3.10), (3.1) holds since applying the natural class extension to $(\;)'$, assuming $\Omega$ boolean here,
$$(a|b)' = (\Omega\cdot b'+a)' = \Omega\cdot b'+a+1 = \Omega\cdot b'+a' = (a'|b).$$

Note next that for all $a,b,c,d \in \Omega$, any commutative ring with unity, again using natural class extension
$$(\Omega\cdot b + a)\cdot(\Omega\cdot d + c) = K(a,b,c,d) + a\cdot b, \quad (3.11)$$
where
$$K(a,b,c,d) \stackrel{d}{=} \{x\cdot y\cdot b\cdot d + y\cdot a\cdot d + x\cdot b\cdot c | x,y \in \Omega\}. \quad (3.12)$$
Also let analogously, again using natural class extension
$$K_0(a,b,c,d) \stackrel{d}{=} \Omega\cdot b\cdot d + \Omega\cdot a\cdot d + \Omega\cdot b\cdot c. \quad (3.13)$$
It follows that letting $x=0$ and then $y=0$ in (3.12), and using similar manipulations that
$$\Omega\cdot a\cdot d \cup \Omega\cdot b\cdot c \subseteq K(a,b,c,d) \subseteq K(a,b,c,d) + \Omega\cdot a\cdot d + \Omega\cdot b\cdot c$$
$$= K(a,b,c,d) + \Omega\cdot b\cdot d$$
$$= K_0(a,b,c,d). \quad (3.14)$$

### Lemma 1.

For all $a,b,c,d \in \Omega$, commutative ring with unity, $K(a,b,c,d)$ is an ideal iff $K(a,b,c,d) = K_0(a,b,c,d)$

iff $\Omega\cdot b\cdot d \subseteq K(a,b,c,d).$ □

Now suppose that $\Omega$ is regular with $a,b,c,d \in \Omega$. Let $z \in \Omega$ be arbitrary and define
$$x_1 \stackrel{d}{=} \lambda_d \cdot d \cdot x_0, \quad (3.15)$$
$$y_1 \stackrel{d}{=} \lambda_b \cdot b \cdot y_0, \quad (3.16)$$
$$x_0 \stackrel{d}{=} 1 - a\cdot\lambda_b, \quad (3.17)$$
$$y_0 \stackrel{d}{=} z - c\cdot\lambda_d + a\cdot\lambda_b\cdot c\cdot\lambda_d. \quad (3.18)$$
Then substituting (3.15),(3.16) into (3.12) yields
$$x_1\cdot y_1\cdot b\cdot d + y_1\cdot a\cdot d + x_1\cdot b\cdot c =$$
$$b\cdot d\cdot(x_0\cdot y_0 + a\cdot\lambda_b\cdot y_0 + c\cdot\lambda_d\cdot x_0)$$
$$=$$
$$b\cdot d\cdot((x_0 + a\cdot\lambda_b)\cdot(y_0 + c\cdot\lambda_d) - a\cdot\lambda_b\cdot c\cdot\lambda_d)$$
$$= b\cdot d\cdot(1\cdot z) = b\cdot d\cdot z, \quad (3.19)$$
which in turn validates the last statement in Lemma 1 and hence by Lemma 1,
$$K(a,b,c,d) = K_0(a,b,c,d), \quad (3.20)$$
Finally, replacing b by b' and d by d' in (3.11), and putting together (3.11),(3.20), and using the property again of regular rings that sums of principal ideals are principal ideals also as computed in (3.10a),
$$(\Omega\cdot b' + a)\cdot(\Omega\cdot d' + c) = \Omega\cdot b'\cdot d' + \Omega\cdot a\cdot d' + \Omega\cdot b'\cdot c + a\cdot c$$
$$= \Omega\cdot(b'\cdot d' \vee a\cdot d' \vee b'\cdot c) + a\cdot c$$
$$= \Omega\cdot((b\vee d)\cdot(a'\vee d)\cdot(b\vee c'))' + a\cdot c$$
$$= \Omega\cdot(a'\cdot b \vee c'\cdot d \vee b\cdot d)' + a\cdot c, \quad (3.21)$$
which is the same as (3.4), using (2.30),(2.31).

Assuming again that $\Omega$ is boolean, since demorgan relations have natural class extensions to $\tilde{\Omega}$, one can use (3.1) as
$$(a|b) \vee (c|d) = ((a|b)'\cdot(c|d)')'$$
$$= ((a'|b)\cdot(c'|d))'$$
$$= (a'\cdot c' | a\cdot b \vee c\cdot d \vee b\cdot d)'$$

$$=(a \lor c | a \cdot b \lor c \cdot d \lor b \cdot d), \quad (3.22)$$

which is the same as (3.3).

The extension of the above results to multiple arguments is tedious and will be omitted. Finally, it should be noted that (3.3) and (3.4) can be extended where for (3.3), $\Omega$ is boolean and for (3.4), $\Omega$ is only regular, where for any ideals $I, J$ of $\Omega$ and all $a, c \in \Omega$,

$$(I + a) \lor (J + c) = (I \cdot J + J \cdot a' + I \cdot c') + (a \lor c), \quad (3.23)$$

$$(I + a) \cdot (J + c) = (I \cdot J + J \cdot a + I \cdot c) + a \cdot c, \quad (3.24)$$

where it should be remarked that in (3.23) and (3.24), on the right hand sides the collections of ideals to the left of $a \lor c$ and $a \cdot c$ are also ideals. Proofs of all of the above results together with other related investigations can be found in [2].

□

Theorem 3.1 specializes, when antecedents are the same, to the formal counterparts of well-used properties of conditional probabilities where the conditioning is upon the same event:

Corollary 3.1

For all $a, b, c \in \Omega$, assumed boolean,

$$(a|b) + (c|b) = (a+c|b) = (a \cdot b + c \cdot b | b), \quad (3.25)$$

$$(a|b) \lor (c|b) = (a \lor c|b) = (a \cdot b \lor c \cdot b | b), \quad (3.26)$$

$$(a|b) \cdot (c|b) = (a \cdot c|b) = (a \cdot b \cdot c | b). \quad (3.27)$$

□

It should be noted that there is a basic compatibility between Corollary 3.1 and the the natural homomorphism $nat_b: \Omega \to \Omega/b'$, where for any $x \in \Omega$,

$$nat_b(x) = (x|b) = x + \Omega \cdot b', \quad (3.28)$$

where all basic properties of $\Omega$ are brought down to the fixed quotient ring $\Omega/b'$ defined through the usual coset operations.

Since all boolean functions over $\Omega$ can be expressed as simple canonical functions of e.g., $\lor, \cdot, ()'$, it follows that the same is true of their natural class extensions and a simple argument thus shows that if $f: \Omega^n \to \Omega$ is any n-ary boolean function, then the natural extension of to $f: \widetilde{\Omega}^n \to \widetilde{\Omega}$ is well-defined.

Returning to the partial order $\leq$ defined over $\Omega$ (boolean, although extendable to regular rings), where

$$a \leq b \text{ iff } a = a \cdot b \text{ iff } b = a \lor b \quad (3.29)$$

and where $\leq$ possesses all the usual lattice properties, it is basic to inquire if the natural class extension of $\leq$ from $\Omega$ to $\widetilde{\Omega}$ preserves these properties.

Theorem 3.2

Let $\Omega$ be boolean. Then define for any $a, b, c, d \in \Omega$

$$(a|b) \leq (c|d) \text{ iff } (a|b) = (a|b) \cdot (c|d). \quad (3.30)$$

Then it follows that

$$(a|b) \leq (c|d) \text{ iff } (c|d) = (a|b) \lor (c|d)$$
$$\text{iff } a \cdot b \leq c \cdot d \text{ and } c' \cdot d \leq a' \cdot b. \quad (3.31)$$

In addition, among the lattice-like properties enjoyed by the legitimate partial order $\leq$ over $\widetilde{\Omega}$ ( since it can be shown to be anti-symmetric, reflexive, and transitive) are, letting $A=(a|b), C=(c|d), E=(e|f), G=(g|h) \in \widetilde{\Omega}$:

$$A \leq C, E \text{ iff } A \leq C \cdot E \text{ ; } C, E \leq A \text{ iff } C \lor E \leq A \text{ ; } \quad (3.32)$$

$$\text{If } A \leq C, \text{ then } C' \leq A'; \quad (3.32a)$$

$$\text{If } A \leq C \text{ and } E \leq G, \text{ then } A \cdot E \leq C \cdot G \text{ and } A \lor E \leq C \lor G. \quad (3.33)$$

Proofs:

The proofs in some cases are rather long, such as for the bottom of (3.31). Again see [2] for detail.

From this point on, most proofs will be omitted and the interested reader is referred to [2].

□

In conjunction with the various properties displayed so far, $\widetilde{\Omega}$ possesses a number of interesting algebraic properties summarized in the next theorem.

Theorem 3.3

Assuming as usual here that $\Omega$ is a boolean ring, $\widetilde{\Omega}$ in general is not a ring due to the failure of additive inverses (-) to hold. However, $\widetilde{\Omega}$ is commutative and associative relative to $+, \lor, \cdot$, has additive identity 0 the same as in $\Omega$ and multiplicative unity 1 the same as in $\Omega$. Also, $\lor$ and $\cdot$ are mutually distributive over $\widetilde{\Omega}$ ; $\widetilde{\Omega}$ (like $\Omega$) is idempotent, demorgan relative to $(\lor, \cdot, ()')$, and $\lor$ and $\cdot$ are mutually absorbing over $\widetilde{\Omega}$ ; finally, $()'$ is involutive over $\widetilde{\Omega}$.

□

Other properties of conditional objects contributing to the development of a calculus of relations are presented below.

Theorem 3.4

For all $a, a_1, \ldots, a_m, b, c, d \in \Omega$, boolean,

$$(a|0) = (0|0) = \Omega, \quad (3.34)$$

$$(1|b) = (b|b) = \Omega \cdot b' + b = \Omega \lor b \quad (3.35)$$

$$(a|b) \cdot (a|b') = (a^2|a') = (a|a') = (0|a') = \Omega \cdot a \quad (3.36)$$

$$(a|b) \lor (a|b') = (a|a), \quad (3.37)$$

$$(a|b) \lor (a|b)' = (b|b), \quad (3.38)$$

$$(a|b) = a + (0|b), \quad (3.39)$$

$$c \lor (a|b) = (a \lor c | b \lor c), \quad c \cdot (a|b) = (c \cdot a | b \lor c'), \quad (3.40)$$

$$c + (a|b) = (c+a|b), \quad (3.41)$$

$$(a|b) + (c|d) = (a|b) \cdot (c|d)' \lor (a|b)' \cdot (c|d), \quad (3.42)$$

$$(a|b \cdot c) \cdot (b|c) = (a \cdot b | c) \text{ (chaining property)}. \quad (3.43)$$

If $a_1, \ldots a_m$ are disjoint and exhaustive, i.e.,

$$a_i \cdot a_j = \delta_{i,j} \text{ and } a_1 + \cdots + a_m = 1, \quad (3.44)$$

then for any $j, j=1, \ldots, m$, the following forms of Bayes' Theorem hold:

$$(a_j|b) = (a_j \cdot b | b) = ( (b|a_j) \cdot a_j | b ), \quad (3.45)$$

$$(a_j|b) \cdot b = (b|a_j) \cdot a_j = a_j \cdot b, \quad (3.46)$$

$$b = (b|a_1) \cdot a_1 + \cdots + (b|a_m) \cdot a_m. \quad (3.47)$$

If $a_1 \leq a_2 \leq \cdots \leq a_m$, then the chaining relation holds:

$$(a_1|a_2) \cdot (a_2|a_3) \cdots (a_{m-1}|a_m) = (a_1|a_m). \quad (3.48)$$

□

The next results tie in conditioning as defined here with classical implication.

Theorem 3.5

For all $a, b \in \Omega$ boolean

$$(a|b) = (b \Rightarrow a|b). \quad (3.49)$$

The smallest element of $(a|b)$ relative to $\leq$ is $a \cdot b$, while the largest element is $(b \Rightarrow a)$, thus

$$a \cdot b \leq (a|b) \leq b \Rightarrow a. \quad (3.50)$$

Also,

275

$$(b \Rightarrow a) = (a|b) \vee b' = (a' \Rightarrow b') = (b'|a') \vee a , \quad (3.51)$$
$$(a|b) = (b \Rightarrow a) \cdot (b|b) = ((b'|a') \vee a) \cdot (b|b), \quad (3.52)$$
$$(b'|a') = (b \Rightarrow a) \cdot (a'|a') = ((a|b) \vee b') \cdot (a'|a'), \quad (3.53)$$
$$(a \Leftrightarrow b) = (a \Rightarrow b) \cdot (b \Rightarrow a) = a \cdot b \vee a' \cdot b' = (a|b) \cdot (b|a) \vee a' b', \quad (3.54)$$
$$(a|b) \cdot (b|a) = (a \cdot b | a \vee b) = (a \Leftrightarrow b) \cdot (a \cdot b | a \cdot b), \quad (3.55)$$

yielding as the smallest element of $(a|b) \cdot (b|a)$ being $a \cdot b$ and the largest being $a \Leftrightarrow b$, analogous to (3.50).

Also, note the pairwise comparisons between $b \Rightarrow a$ and $(a|b)$: (See also Table 1, Section 1.)

$$(a|b) = (a \cdot b | b) \text{ while } (b \Rightarrow a) = (b \Rightarrow a \cdot b), \quad (3.56)$$
$$(1|b) = (b|b) = \Omega \vee b \text{ while } (b \Rightarrow 1) = (b \Rightarrow b) = 1, \quad (3.57)$$
$$(b|1) = b \text{ while } (1 \Rightarrow b) = b, \quad (3.58)$$
$$(b|0) = \Omega \text{ while } (0 \Rightarrow b) = 1, \quad (3.59)$$
$$(a|b)' = (a' \cdot b | b) \text{ while } (b \Rightarrow a)' = a' \cdot b, \quad (3.60)$$
$$(0|b) = (b'|b) = \Omega \cdot b' \text{ while } (b \Rightarrow 0) = (b \Rightarrow b') = b', \quad (3.61)$$
$$(a|b) \cdot (c|d) = (a \cdot c | q) \text{ while } (b \Rightarrow a) \cdot (d \Rightarrow c) = (q \Rightarrow ac), \quad (3.62)$$

where
$$q \triangleq a' \cdot b \vee c' \cdot d \vee b \cdot d. \quad (3.63)$$

Next,
$$(a|b) \vee (c|d) = (a \vee c | r) \text{ while } (b \Rightarrow a) \vee (d \Rightarrow c) = (b \cdot d \Rightarrow (a \vee c)), \quad (3.64)$$

where
$$r \triangleq a \cdot b \vee c \cdot d \vee b \cdot d. \quad (3.65)$$

Also, for $a \le b \le c$, transitivity holds as
$$(a|b) \cdot (b|c) = (a|c) \text{ while } (c \Rightarrow b) \cdot (b \Rightarrow a) \le (c \Rightarrow a), \quad (3.66)$$

and for $a \le b \cdot c$, improvement of information is
$$(a|b) \le (a|b \cdot c) \text{ while } (b \Rightarrow a) \le (b \cdot c \Rightarrow a). \quad (3.67)$$

Also, referring to Section 4, and the class reduction operator $\bar{u}$, one can compare iterated classical implication and iterated conditional forms
$$\bar{u}((a|b)|(c|d)) = (a|\alpha) \text{ while } ((d \Rightarrow c) \Rightarrow (b \Rightarrow a)) = (\gamma \Rightarrow a), \quad (3.68)$$

where
$$\alpha = b \cdot (c \cdot d \vee a' \cdot d'), \quad \gamma = b \cdot (c \cdot d \vee d'), \quad (3.69)$$

with the special cases
$$\bar{u}((a|b)|(c|b)) = \bar{u}((a|b)|c)) = (a|b \cdot c) \text{ while}$$
$$((b \Rightarrow c) \Rightarrow (b \Rightarrow a)) = (c \Rightarrow (b \Rightarrow a)) = ((b \cdot c) \Rightarrow a). \quad (3.70)$$
□

Finally, this section is concluded with a result which is not only interesting in its own right as a generalization of the classic result concerning the disjointness or identity of cosets having the same antecedent, but which is useful in further analysis of conditional objects.

### Theorem 3.6

For any $a,b,c,d \in \Omega$ boolean and denoting $\cap$ below for the ordinary class intersection,

$$(a|b) \cap (c|d) = \begin{cases} \emptyset & \text{iff } a+c \notin (0|b \cdot d) \\ (\xi | b \vee d) & \text{iff } a+c \in (0|b \cdot d), \end{cases} \quad (3.71)$$

where
$$\xi \triangleq p + a = q + c ; a + c = q + p, \quad (3.72)$$
for some $p \in (0|b)$ and $q \in (0|d)$.

From the above it follows that
$$(a|b) \subseteq (c|d) \text{ iff } (d \le b \text{ and } a \in (c|d)). \quad (3.73)$$
□

## 4. ADDITIONAL PROPERTIES OF CONDITIONAL OBJECTS

In the last section, a basic calculus of operations was presented for conditional objects. In this section, certain selected topics involving conditioning are briefly considered.

First, define higher order conditional objects through natural class extensions of conditional objects as defined in the previous sections. Thus for any $a,b,c,d \in \Omega$ boolean, define

$$((a|b)|(c|d)) = f^{-1}_{(c|d)}((a|b) \cdot (c|d))$$
$$= \{(x|y) | (x|y) \in \tilde{\Omega} \text{ and } (x|y) \cdot (c|d) = (a|b) \cdot (c|d)\}. \quad (4.1)$$

Some basic properties and an explicit solution are given next.

### Theorem 4.1

For all $a,b,c,d \in \Omega$ boolean:
Analogous to (2.20),
$$((a|b)|(c|d)) = ((a|b) \cdot (c|d) | (c|d)) \in P(\tilde{\Omega}) \quad (4.2)$$
and
$$((a|b)|(c|d)) \cdot (c|d) = (a|b) \cdot (c|d). \quad (4.3)$$

Without loss of generality, using (2.20),(3.31), and (4.2), assume from now on, unless otherwise stated:
$$a \le b , c \le d , (a|b) \le (c|d). \quad (4.4)$$

Then explicitly
$$((a|b)|(c|d)) = (a|b) \vee T_{\beta,c,d}$$
$$= (a|b) \vee \beta \cdot V_{c,d}$$
$$= \{\tau_{a,b,\beta;s,t} | t \le s \in \Omega\}, \quad (4.5)$$

where for all $t \le s \in \Omega$,
$$\tau_{a,b,\beta;s,t} \triangleq (a|b) \vee t_{\beta,c,d;s,t}$$
$$= (a \vee \beta \cdot t | a \vee \beta \cdot t \vee b \cdot ((c' \cdot d)' \vee s))$$
$$= (a \vee \beta \cdot t | a \vee \beta \cdot t \vee b \cdot (\beta' \vee s)) \in \Omega, \quad (4.6)$$
$$\beta \triangleq b' \cdot d' \vee c' \cdot d, \quad (4.7)$$

resulting in
$$\beta' = (b \vee d) \cdot (c' \cdot d)' = c \vee b \cdot d', \quad (4.8)$$
$$T_{\beta,c,d} \triangleq \{t_{\beta,c,d;s,t} | t \le s \in \Omega\} = \beta \cdot V_{c,d}, \quad (4.9)$$
$$t_{\beta,c,d;s,t} \triangleq \beta \cdot \delta_{c,d;s,t}$$
$$= (\beta \cdot t | (c' \cdot d)' \vee s) \in \Omega, \quad (4.10)$$
$$\delta_{c,d;s,t} \triangleq (t | (c' \cdot d)' \vee s) \in \Omega, \quad (4.11)$$
$$V_{c,d} \triangleq \{\delta_{c,d;s,t} | t \le s \in \Omega \}. \quad (4.12)$$
□

Unfortunately, unlike the single conditional case (see (2.22)), second level conditional objects pose a problem with respect to both uniqueness of representations relative to their antecedents and the closure of boolean operations. Surprisingly, only three parameter values - out of four possible a priori - are required to specify such forms uniquely. The representation is characterized in the following theorem:

### Theorem 4.2

For all $a_i,b_i,c_i,d_i \in \Omega$, boolean satisfying (4.4) ($a$ replaced by $a_i$, etc.) and $\beta_i, \beta_i'$ as in (4.7),(4.8), respectively (with $b$ replaced by $b_i$, etc.), $i=1,2$:
$$((a_1|b_1)|(c_1|d_1)) = ((a_2|b_2)|(c_2|d_2)) \text{ iff } (a_1=a_2, b_1=b_2, \beta_1=\beta_2)$$
$$\text{iff } (a_1=a_2, b_1=b_2, c_2=a_1 \vee \beta_1' \cdot (b_1' \vee w), d_2=c_2 \vee \beta_1 \cdot b_1), \quad (4.13)$$
for any fixed $w \in \Omega$. □

276

One reasonable way to treat the difficulties arising from the necessary introduction of iterated conditional forms is to determine if there is some mapping from these higher levels down to the single level which can be used to identify the former with the latter.

As a candidate for the above, suppose we consider the class reduction operator $\bar{u}:P(P(\Omega)) \to P(\Omega)$, where for all $A \in P(P(\Omega))$,

$$\bar{u}(A) \overset{d}{=} \bigcup_{A \in A} A = \{x | x \in A \in A\} \subseteq \Omega . \quad (4.14)$$

**Theorem 4.3**

Let $\Omega$ be boolean and denote, analogous to $\tilde{\Omega}$ being the class of all (single) conditional objects formed from $\Omega$, $\tilde{\tilde{\Omega}}$ as the class of all (double) conditional objects formed from $\tilde{\Omega}$, noting that $\Omega \subseteq \tilde{\Omega} \subseteq \tilde{\tilde{\Omega}}$. Then:

$\bar{u}:\tilde{\tilde{\Omega}} \to \tilde{\Omega}$ is a surjective homomorphism relative to all boolean operations extended in a natural class way from $\Omega$.

Furthermore, the specific relation defining $\bar{u}$ can be determined to be, for all $a,b,c,d \in \Omega$ satisfying (4.4),

$$\bar{u}((a|b)|(c|d)) = (a|b \cdot \beta') = (a|b \cdot (c' \cdot d)') . (4.15)$$

In particular,

$$\bar{u}(a|b) = (a|b), \quad (4.16)$$

$$\bar{u}((a|b)|(c|b)) = \bar{u}((a|b)|c) = (a|b \cdot c), \quad (4.17)$$

$$\bar{u}(a|(c|d)) = (a|(c' \cdot d)'), \quad (4.18)$$

$$(c|d) \cdot \bar{u}((a|b)|(c|d)) = (a|b). \quad (4.19)$$

Also, the following restrictions of $\bar{u}$ are surjective isomorphisms relative to all boolean operations extended in a natural class way:

$$\bar{u}: \{((a|b)|c) \mid a,b \in \Omega\} \to \{(a|b \cdot c) \mid a,b \in \Omega\}, (4.20)$$

$$\bar{u}: \{(a|b)|(c|b) \mid a,c \in \Omega\} \to \{(a|b \cdot c) \mid a,c \in \Omega\}, (4.21)$$

$$\bar{u}: \{(a|b)|(c|d) \mid a,b \in \Omega\} \to \{(a|b \cdot \beta') \mid a,b \in \Omega\}, (4.22)$$
□

Thus, in a natural way, one can identify all higher order conditional objects with single conditional ones.

Finally, we conclude this paper with some results involving conditional objects and conditional probabilities directly.

Firstly, recall that conditional probabilities can be considered a homomorphic evaluation of the formal relations in (2.26) (left side) (see also (1.2)), as well as (3.1) and (3.25)-(3.27). (Again, see Theorem 2.1.) Also, conditional probabilities can be identified, with the introduction of conditional objects, as the extension of probability measure $p:\Omega \to [0,1]$ to monotone function $p:\tilde{\Omega} \to [0,1]$, i.e., if $(a|b) \leq (c|d) \in \tilde{\Omega}$, then

$$p((a|b)) = p(a|b) \leq p(c|d) = p((c|d)). \quad (4.23)$$

In particular, this shows that (3.50) implies, as a check, (1.4). Other inequalities can be similarly established through first using the formal counterparts. One can also define measure-free independence of conditional objects $(a|b)$ and $(c|d)$ to occur when they are p-independent, i.e,

$$p((a|b) \cdot (c|d)) = p(a|b) \cdot p(c|d), \quad (4.24)$$

for all possible probability measures $p:\Omega \to [0,1]$. One simple example of such a measure-free pair is $(a|b), b$, using (2.26)(left side). This can be extended to certain other pairs and to multiple conditional forms as the factors in (3.48). (See [1], Section 5.) Also sequential updating of information can be very elegantly described through the use of conditional objects rather than only through conditional probabilities ([1], Section 5).

Lastly, we consider briefly random conditional objects and how they relate to conditional probabilities. Beginning with probability space $(M,A,p)$ and r.v. $V:M \to R^m$, r.v. $W:M \to R^n$, extend $V,W,V \times W$ in the natural class sense to $V:\tilde{A} \to \tilde{B}^m$, $W:\tilde{A} \to \tilde{B}^n$, $V \times W:\tilde{A} \to \tilde{B}^{m+n}$, respectively, where for all $a,b \in A$,

$$V(a|b)=V(a) \times R^n; W(a|b)=R^m \times W(b); (V \times W)(a|b)=V(a) \times W(b). (4.25)$$

Then the *random conditional object* $(V|W): \tilde{A} \to \tilde{B}^{m+n}$ is defined by, for all $a,b \in A$,

$$(V|W)(a|b) \overset{d}{=} ((V \times W)(a|b)|W(a|b)) = (V(a) \times W(b)|R^m \times W(b))$$
$$\overset{d}{=} (V(a)|W(b)), \quad (4.26)$$

with inverse mapping $(V|W)^{-1}:\tilde{B}^{m+n} \to A$, yielding for any $c \in B^m$, $d \in B^n$,

$$(V|W)^{-1}(c|d)=((V \times W)^{-1}(c \times d)|W^{-1}(d))=(V^{-1}(c)|W^{-1}(d)). (4.27)$$

Thus, $(V|W)$ induces "conditional event probability space" ($R^{m+n}, \tilde{B}^{m+n}, P_{(V|W)}$), where $P_{(V|W)}: \tilde{B}^{m+n} \to [0,1]$ is given by

$$P_{(V|W)}(c|d) \overset{d}{=} p((V|W)^{-1}(c|d)) = p(V^{-1}(c)|W^{-1}(d)). \quad (4.28)$$

By using an optimal approximation technique, arithmetic operations over conditional objects can also be determined, in turn yielding expectations of random conditional objects, defined in the natural way. Thus, e.g,

$$E((V|W)) = ( E(V \times W) | E(V \times W) ), \quad (4.29)$$

where $E(\cdot)$ is ordinary expectation. (See [1], Section 5.)